\documentclass{article}

\usepackage{PRIMEarxiv}

\usepackage[utf8]{inputenc} % allow utf-8 input
\usepackage[T1]{fontenc}    % use 8-bit T1 fonts
\usepackage{hyperref}       % hyperlinks
\usepackage{url}            % simple URL typesetting
\usepackage{booktabs}       % professional-quality tables
\usepackage{amsfonts}       % blackboard math symbols
\usepackage{nicefrac}       % compact symbols for 1/2, etc.
\usepackage{microtype}      % microtypography
\usepackage{lipsum}
\usepackage{fancyhdr}       % header
\usepackage{graphicx}       % graphics
\graphicspath{{media/}}     % organize your images and other figures under media/ folder
\usepackage{mathtools,amsfonts}
\DeclareMathOperator{\E}{\mathbb{E}}
\title{High Fidelity Synthetic Face Generation for Rosacea Skin Condition from Limited Data
}

\author{
Anwesha Mohanty \\
Dublin City University \\
\texttt{anwesha.mohanty2@mail.dcu.ie} \\
\And
Alistair Sutherland \\
Dublin City University \\
\texttt{alistair.sutherland@dcu.ie} \\
\And
Marija Bezbradica \\
Dublin City University \\
\texttt{marija.bezbradica@dcu.ie} \\
\And
Hossein Javidnia \\
Dublin City University \\
\texttt{hossein.javidnia@dcu.ie} \\
}

\begin{document}
\maketitle

\begin{abstract}
Similar to the majority of deep learning applications, diagnosing skin diseases using computer vision and deep learning often requires a large volume of data. However, obtaining sufficient data for particular types of facial skin conditions can be difficult due to privacy concerns. As a result, conditions like Rosacea are often understudied in computer-aided diagnosis. The limited availability of data for facial skin conditions has led to the investigation of alternative methods for computer-aided diagnosis. In recent years, Generative Adversarial Networks (GANs), mainly variants of StyleGANs, have demonstrated promising results in generating synthetic facial images. In this study, for the first time, a small dataset of Rosacea with 300 full-face images is utilized to further investigate the possibility of generating synthetic data. The preliminary experiments show how fine-tuning the model and varying experimental settings significantly affect the fidelity of the Rosacea features. It is demonstrated that $R_1$ Regularization strength helps achieve high-fidelity details. Additionally, this study presents qualitative evaluations of synthetic/generated faces by expert dermatologists and non-specialist participants. The quantitative evaluation is presented using a few validation metric(s). Furthermore a number of limitations and future directions are discussed. Code and generated dataset are available at: \url{https://github.com/thinkercache/stylegan2-ada-pytorch}
\end{abstract}

% keywords can be removed
\keywords{Limited Data \and Generative Adversarial Networks \and Regularization \and Dermatology \and Synthetic Image Generation \and Computer-aided diagnosis}

\section{Introduction}
\label{sec:Introduction}
Computer-aided diagnosis of skin diseases has become more popular since the introduction of Inception v3 \cite{szegedy2016rethinking} that achieved a performance accuracy of  93.3\% \cite{esteva2017dermatologist} in classifying various cancerous skin conditions. A large dataset with approximately 129,450 images has been utilized to develop the skin cancer classification model with Inception v3\cite{szegedy2016rethinking}. However, gathering such a large amount of data is not feasible for some skin conditions such as Rosacea. Although many skin conditions can lead to fatal consequences, cancer has been considered the most serious of all and has motivated the gathering of more data over time. As a result, many Teledermatology \cite{pala2020teledermatology} websites have a substantial amount of skin cancer images. On the other hand, there is very limited data for  non-fatal chronic skin conditions such as Rosacea. Deep Convolutional Neural Networks (DCNNs) e.g. Inception v3, perform relatively well provided with a large training dataset \cite{najafabadi2015deep}. However, their performance significantly degrades in the absence of large amounts of data. A possible solution is to utilize the small amount of available data by leveraging the idea of Generative Adversarial Networks (GANs)\cite{goodfellow2020generative} to produce synthetic images. Synthetic images may help to expand the existing small dataset by an adequate amount which may help to train a DCNN. Generating synthetic datasets for diseases may also help educate non-specialist populations to create awareness and advertisement. This research aims to generate synthetic images by means of expanding a small dataset for Rosacea skin condition using a variant of the StyleGAN architecture\cite{karras2019style} trained with only 300 images of Rosacea.

Although, there have been a few studies \cite{baur2018melanogans, bissoto2018skin, pollastri2020augmenting, ghorbani2020dermgan, fossen2020synthesizing, bissoto2021gan} on generating synthetic images of skin cancer lesions using various types of GANs architectures, the images are captured through a dermatoscope and other imaging devices that focus only on a specific locality i.e. cancerous region of the skin. In contrast, the Rosacea dataset used in our study contains full-face images. Hence, the modalities of skin cancer images and full-face images with Rosacea are entirely different. An important reason to consider a full-face image for Rosacea analysis is that different subtypes of the disease can affect multiple parts of the face. The impact of Rosacea on facial skin can be assessed by considering different local regions of the skin and diagnosing the subtype of Rosacea. 

\subsection{A brief introduction to the skin diseases and Rosacea}
\label{sec:A brief introduction to the skin diseases and Rosacea}
The observational and analytical complexities of skin diseases are challenging aspects of diagnosis and treatment. In most cases, at the early stage, skin diseases are examined visually. Depending on the complexity of the early examination and severity of the disease, a few clinical or pathological measures using images of the affected region may be followed. These include dermoscopic analysis, biopsy, and histopathological examination. Depending on the nature of the skin disease, whether it is acute or chronic, the diagnosis and treatment may be time-consuming. 

In this work, we focus on a chronic inflammatory skin disease called Rosacea. Rosacea is a chronic facial skin condition and a cutaneous vascular disorder that goes through a cycle of fading and relapse\cite{del2012Rosacea}. It is a common skin condition in native people from northern countries with fair skin or with Celtic origins \cite{powell2005Rosacea}. Rosacea is often characterised by signs of facial flushing and redness, inflammatory papules, and pustules, telangiectasias, and facial edema. The symptom severity varies greatly among individuals\cite{steinhoff2013new}. In the medical diagnostic approach, Rosacea is classified into four subtypes – Subtype 1 (Erythematotelangiectatic Rosacea), Subtype 2 (Papulopustular Rosacea), Subtype 3 (Phymatous Rosacea) and Subtype 4 (Ocular Rosacea). Each subtype may be further classified based on the severity of the condition, e.g., mild, moderate, or severe \cite{johnston2018experiences}. In this work, we are considering subtype 1 and subtype 2.

In most situations, Rosacea is not fatal but may still need lifelong engagement with dermatologists. Because it is a chronic condition, Rosacea requires regular check-ups, up-to-date medications, and surgical or laser treatments, if necessary. This routine diagnosis and treatment require a significant amount of time and cost, starting from getting an appointment with a dermatologist to getting treated. To prevent further complications in the condition, diagnosis at the early stages of the disease allows for intervention and prevention.
Since the early 90s, digital platforms have been helping dermatologists to discuss and to diagnose skin diseases. This collaborative practice has been utilising skin disease images and additional health data from patients around the globe. In the medical literature, this procedure of collecting, monitoring, storing, and sharing data to help diagnose skin conditions is termed “Teledermatology”\cite{pala2020teledermatology}. Teledermatology has been providing a great channel to explore the field of computer-aided diagnosis by using medical/clinical images of skin diseases through advanced computer vision and machine learning techniques.

\subsection{Importance of studying full-face images of Rosacea}
\label{sec:Importance of studying full-face images of Rosacea}
Generally, prolonged redness is one of the common early symptoms (pre-Rosacea) that usually appears over the cheeks, chin, nose, or forehead. Eventually, certain patients develop some swelling (‘edema’ in medical terminology), which is noticeable at the very early stage of the disease. Particularly, the locality and the visible regions of blood vessels (‘telangiectasia’ in medical terminology), prolonged redness and edema may give the impression of the severity of the disease. Full-face images of patients have proved very useful in diagnosing the condition and predicting the timeline of growth for future diagnosis and treatment. Hence, in this study, we are considering full-face images of Rosacea, while most of the previous studies mentioned in \ref{sec:Related work on Rosacea diagnosis} have considered only partial images of the face.

\subsection{Limited data availability for Rosacea}
\label{sec:Limited data availability for Rosacea}
There are only a few hundred images publicly available for analysis and diagnosis of Rosacea \cite{mohanty2022skin}. Among the available images, only a small number have full-face visibility. Some of the datasets with full-face visibility are watermarked, which does not satisfy our selection criteria, as discussed further in Section \ref{sec:Publicly available data} and \ref{sec:Rosacea Dataset- ‘rff-300’}. There are a few teledermatology websites which have images of Rosacea available publicly for research namely Dermatology ATLAS \cite{dermatologyatlasbrazil}, DanDerm \cite{dermatologyatlasdenmark}, DermIS \cite{dermis}, DermNetNZ \cite{dermnetnz}, Dermatoweb.net \cite{dermatowebspain}, Hellenic Dermatological Atlas \cite{hellenic}.

Compared to recent studies published on skin cancer classification, which all use an adequate number of images, there is a very limited number of annotated Rosacea images. This introduces a significant challenge for the dataset split (train, validation, and test) needed when training deep learning models.

\subsection{Related work on Rosacea diagnosis}
\label{sec:Related work on Rosacea diagnosis}
There have been a few noteworthy works conducted on Rosacea by Thomsen et al.\cite{thomsen2020deep}, Zhao et al.\cite{zhao2021novel}, Zhu et al.\cite{zhu2021deep}, Binol et al.\cite{binol2020ros}, and Xie et al.\cite{xie2019xiangyaderm} with significant quantities of data collected from dermatology departments in hospitals. However, the datasets used in these studies were entirely confidential. In these studies, the early detection problem of Rosacea is addressed by performing ‘image classification’ among different subtypes of Rosacea and other common skin conditions. The classifier was trained using data augmentation and transfer learning from the pre-trained weights of ImageNet. In total, there are over 10,000 images used in these studies, along with transfer learning. Transfer learning works well with a significant number of images available, typically over 1000. Following the previous studies mentioned, Mohanty et al.\cite{mohanty2022skin} conducted several experiments on full face Rosacea image classification using Inception v3\cite{szegedy2016rethinking} and VGG16\cite{simonyan2014very}. In the experiments, the aforementioned deep learning models tend to overfit during training and validation due to insufficient data. The main research gap in computer-aided Rosacea diagnosis is the access to sufficient amount of rosacea images for analysis and classification. This motivated exploring the techniques by which small datasets can be leveraged for specific disease categories like Rosacea. Therefore, this work explores the potentials of GANs in generating a synthetic dataset of full faces with Rosacea given a scarce/small dataset. This experimentation may prove helpful in the low data regime in medical image analysis. 

To summarize, our \textbf{contributions} are as follows:

\begin{enumerate}
    \item In this study, to the best of our knowledge, for the first time, a small dataset of Rosacea with 300 full-face images is utilized for synthetic image generation.  
    
    \item We show how fine-tuning the model (StyleGAN2-ADA) and varying experimental settings significantly affect the fidelity of Rosacea features. 
    
    \item We demonstrate that $R_1$ Regularization strength helps achieve high-fidelity characteristics of Rosacea condition. 
    
    \item We generate 300 high-fidelity synthetic full-face images with Rosacea, which can be further utilized to expand the Rosacea face dataset for computer-aided clinical diagnosis. 
    
    \item We present qualitative evaluations of synthetic/generated faces by expert dermatologists and non-specialist participants, these show the realistic characteristics of Rosacea in generated images. 
    
    \item We critically analyse the quantitative evaluation such as validation metrics(s) from list of conducted experiments and point out the limitations of usage of validation metric(s) alone as evaluation criteria in computer-aided medical image diagnosis field. 
\end{enumerate}

\section{Background and related work on synthetic facial image generation}
\label{sec:Background and related work on synthetic facial image generation}
The very first facial image generator using \textit{Generative Adversarial Networks (GANs)} was designed by Goodfellow et al.\cite{goodfellow2020generative} in 2014. The generated synthetic faces were very noisy and required more work to make them convincing. Later, in 2015, \textit{Deep Convolutional GANs (DCGANs)}\cite{radford2015unsupervised} were introduced and used 350,000 face images without any augmentation. DCGANs came with some notable features that resulted in better synthetic faces, such as: 
\begin{itemize}
    \item Improved architectural topology,
    \item Trained discriminators, 
    \item Visualization of filters, 
    \item Generator manipulation. 
\end{itemize}
However, DCGANs model came with some limitations noticeable in : 
\begin{itemize}
    \item Model instability, 
    \item Mode collapse, 
    \item Filter leakage after a longer training time,
    \item Small resolutions of generated images.  
\end{itemize}
These limitations strongly influenced the topics for future work on GANs.

The \textit{Progressive Growing of GANs (ProGANs)} introduced by Karras et al.\cite{karras2017progressive}, improved the resolution of the generated images with a stable and swifter training process. The main idea of ProGANs is to start from a low resolution i.e. $4\times4$ and then progressively increase the resolution, e.g. up to $1024\times1024$, by adding layers to the networks. The training time is 2-6 times faster depending on the desired output resolution. ProGANs could generate $1024\times1024$ facial images using CelebA-HQ\cite{karras2017progressive} dataset with 30,000 selected real images in total. The idea of ProGAN emerged from one of the GANs architectures introduced by Wang et al.\cite{wang2018high}. Although ProGAN successfully generated facial images with large resolution, it did not work adequately in generating realistic features and microstructures. 

\begin{figure*}[t!]
\centering
\includegraphics[width=1\textwidth, angle=0]{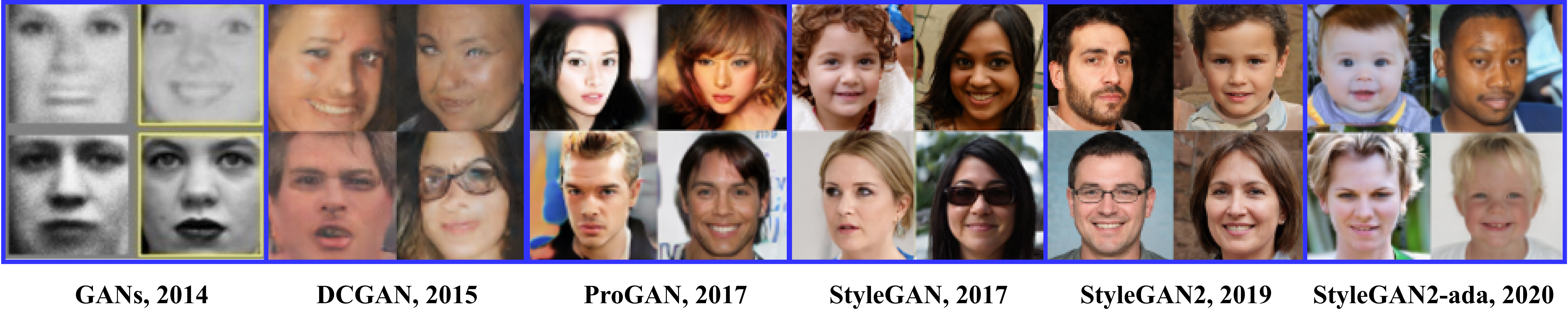}
\caption{Progress of synthetic face generation using various GAN models with the maximum volume of dataset available.}
\label{fig1}
\end{figure*}

\par
Although the generation of high-resolution images was achieved by GANs, there were still indispensable research gaps that needed to be addressed. Thus, the introduction of \textit{StyleGAN}\cite{karras2019style} came with further improvements which helped in understanding various characteristics and phases in synthetic image generation/image synthesis. Important improvements in the StyleGAN architecture include:
\begin{itemize}
    \item Upgrading the number of trainable parameters in style-based generators; this is now 26.2 million, compared to 23.1 million parameters in the ProGAN\cite{karras2017progressive} architecture. 
    \item Upgrading the baseline using upsampling and downsampling operations, increasing training time and tuning the hyperparameters.
    \item Adding a mapping network and adaptive instance normalization (AdaIN) operations. 
    \item Removing the traditional input layer and starting from a learned constant tensor which is $4\times4\times512$.
    \item Adding explicit uncorrelated Gaussian noise inputs, which improves the generator by generating stochastic details. 
    \item Mixing regularization which helps in decorrelating the neighbouring styles and taking control of fine-grained details in the synthetic images. 
\end{itemize}

In addition to the improvements in generating high-fidelity images, StyleGAN introduced a new dataset of human faces called Flickr Faces HQ (FFHQ). FFHQ has 70,000 images at 1024 $\times$ 1024 resolution and is of a diverse range of ethnicity, age, background artifacts, make-up, lighting, image viewpoint and various accessories such as eyeglasses, hats, sunglasses etc. Based on these improvements, comparative outcomes are evaluated using a metric called Fréchet Inception Distance (FID)\cite{heusel2017gans} on two datasets i.e. CelebA-HQ\cite{karras2017progressive} and FFHQ. The recommended future investigations include separating high-level attributes and stochastic effects while achieving linearity of the intermediate latent space.

\begin{figure}[htbp]
\centerline{\includegraphics[width=1\textwidth, angle=0 ]{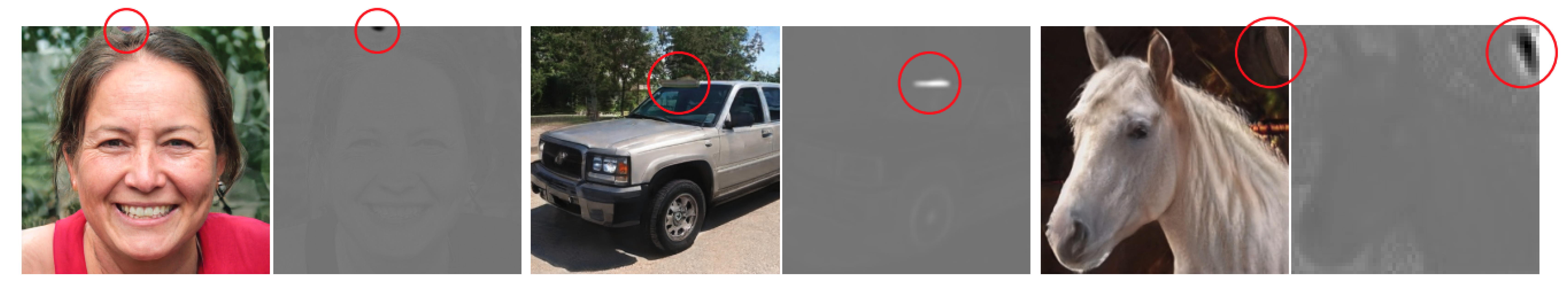}}
\caption{An example of blob-like artifacts in the generated images. This image is taken from Karras et al.\cite{karras2020analyzing}}
\label{fig2}
\end{figure}

Successively, another variant of StyleGAN was introduced by Karras et al. called \textit{StyleGAN2}\cite{karras2020analyzing}, in which the key focus was exclusively on the analysis of the latent space $W$. As the generated output images from StyleGAN contained some unnecessary and common blob-like artifacts, StyleGAN2 addressed the causes of these artifacts and eliminated them by defining some changes in the generator network architecture and in the training methods. Hence the generator normalization is redesigned, and the generator regularization is redefined to boost conditioning and to improve output image quality. The notable improvements in the StyleGAN2 architecture include:
\begin{itemize}
    \item The presence of blob-like artifacts such as those in Fig. \ref{fig2} solved by removing the normalization step from the generator (Generator redesign).
    \item Grouped convolutions are employed as a part of Weight demodulation, in which weights and activation functions are temporarily reshaped. In this setting, one convolution sees one sample with $N$ groups, instead of $N$ samples with one group.  
    \item Adaption of \textit{Lazy Regularization} in which $R_1$ regularization is performed only once in 16 mini-batches. This reduces computational costs and memory usage in total. 
    \item Adding a path length regularization aids in model reliability and performance. This offers a wide scope for exploring the architecture at the further stage. Path length regularization helps in creating denser distributions without mode collapse problems. 
    \item Revisiting the ProGAN architecture to adapt benefits and remove the drawbacks e.g. progressive growing in the residual block of the discriminator network. 
\end{itemize}

The datasets LSUN\cite{yu2015lsun} and FFHQ were used with StyleGAN2 to obtain  quantitative results through metrics such as FID\cite{heusel2017gans}, Perceptual Path Length (PPL)\cite{karras2019style}, Precision and Recall\cite{kynkaanniemi2019improved}.

Another set of GAN architectures called \textit{BigGAN} and \textit{BigGAN-deep}\cite{brock2018large} expanded the variety and fidelity of the generated images. These improvements included making architectural changes which improved scalability, a regularization scheme to recuperate conditioning as well as to boost performance. The above modifications gave a lot of freedom to apply the \textit{“truncation trick”}, a sampling method that aids in controlling sample variety and fidelity in the image generation stage. Even though different GAN architectures produced improved results over a period, model instability during training was a common problem in large-scale GAN architectures\cite{lucic2018gans}. This problem was investigated and analysed through the introduction of BigGAN by leveraging the existing techniques and by presenting novel techniques. The ImageNet ILSVRC 2012 dataset\cite{russakovsky2015imagenet} with the resolutions $128\times128, 256\times256, 512\times512$ was used in BigGAN and BigGAN-deep architectures for demonstrating quantitative results through metrics such as FID and Inception Score (IS)\cite{salimans2016improved}. 

The aforementioned GAN architectures were trained on a large amount of data and can generate high-resolution outputs with variety and a fine-grained texture. Although a large amount of data helps GAN models to learn and generate more realistic-looking synthetic images, it is not possible to acquire a large amount of data for certain fields/domains. For example, in the medical/clinical imaging domain, it is hard to acquire a large number of images for each disease case. Therefore, it is important to expand the potential of GAN architectures to perform well and produce high-fidelity synthetic images, even if there are limited images available. 

However, the key problem with a small number of images is the overfitting of training examples in the discriminator network. Hence the training process starts to diverge, and the generator does not generate anything meaningful because of overfitting. The most common strategy to tackle overfitting in deep learning models is \textit{“data augmentation”}. There are instances in which augmentation functions learn to generate the augmented distribution, which results in \textit{“leaking augmentations”} in the generated samples. The leaking augmentations are the features which are learned from the augmentation style rather than the features which are originally present in the real dataset. 

Hence to prevent the discriminator from overfitting when there is only limited data available, a variant of StyleGAN2 called \textit{StyleGAN2-ADA}\cite{karras2020training} has been introduced with a wide range of augmentations. An adaptive control scheme was presented in order to prevent such augmentations from leaking in the generated images. This work produced promising results in generating high resolution synthetic images obtained with a few thousand images. The significant improvements in StyleGAN2-ADA include:
\begin{itemize}
    \item \textit{Stochastic Discriminator Augmentation} is a flexible method for augmentation that prevents the discriminator from becoming overly confident by showing all the applied augmentation to the discriminator. This assists in generating desired outcomes.
    \item Addition of \textit{Adaptive Discriminator Augmentation (ADA)} by which the strength of augmentation \textit{‘p’} can be adjusted at every interval of 4 mini-batches $N$. This technique helps in achieving convergence during training without any occurrence of overfitting irrespective of the volume of the input dataset.
    \item \textit{Invertible transformations} are applied to leverage the full benefit of the augmentation. The proposed augmentation pipeline contains 18 transformations grouped in 6 categories viz. pixel blitting, more general geometric transformations, colour transforms, image-space filtering, additive noise, and cutout. 
    \item Capability to handle small-volume datasets such as, 1000 and 2000 images from FFHQ dataset, 1336 images of METFACES \cite{metfaces}, 1994 overlapping cropped images from 162 breast cancer histopathology images called BRECAHAD\cite{aksac2019brecahad}, nearly 5,000 images of AFHQ and 50,000 images of CIFAR-10 \cite{krizhevsky2009learning}. 
    \item Although the small volume of the dataset is the main feature in the StyleGAN2-ADA, some high-volume datasets are broken down into different sizes for monitoring the model performance. The FFHQ dataset is used for training the model. Various subsets of the dataset such as 140,000, 70,000, 30,000, 10,000, 5000, 2000 and 1000 are used to test the performance. Similarly, dataset LSUN CAT is considered with the volume starting from 200k to 1k for model evaluation. The FID is used as an evaluation metric for comparative analysis and demonstration of StyleGAN2-ADA model performance.
\end{itemize}

Amongst the studies and related work regarding face generation using GANs as discussed above and represented in Fig. \ref{fig1}, StyleGAN2-ADA appeared to work adequately with a small volume of data. Especially in the case of small volumes of medical/clinical images, StyleGAN2-ADA is a useful method for investigation. Considering the advantages of StyleGAN2-ADA, in this research, we implemented and trained the model with 300 images of Rosacea, to be discussed in section \ref{sec:Experiments and Results}.

\section{Methodology}
\label{sec:Methodology}
\subsection{StyleGAN2 with Adaptive Discriminator Augmentation}
\label{sec:StyleGAN2 with Adaptive Discriminator Augmentation}
The above analysis of the state-of-the-art techniques indicates that StyleGAN2-ADA can potentially be used to address the data scarcity of Rosacea by generating synthetic samples.

The most attractive point of StyleGAN2-ADA is its ability to handle a small amount of data, in fact a minimum of 1000 images. This is achieved by utilizing the concept of \textit{Adaptive Discriminator Augmentation (ADA)}. 

The concept of \textit{ADA} is motivated by three well-known \textbf{limitations} of GAN models \cite{goodfellow2020generative}\cite{gui2021review}:   
\begin{enumerate}
    \item Difficulty in handling small amounts of data.
    \item Discriminator overfitting which leads to mode collapse. 
    \item Sensitivity to the selection of hyperparameters. 
\end{enumerate}

Generally, when condition \textit{1} exists, it is more probable for condition \textit{2} to occur, and when both exist, it leads to catastrophic failure in most GAN models. Nevertheless, when limited data is available, one possible solution for overfitting is “Data Augmentation”. Data Augmentation helps in expanding the input images by applying temporary alterations such as geometric transformations and preprocessing tasks. This practice helps in increasing input feature space during the training. 

However, these augmentations can have negative effects, as most of the existing GAN models augment the real images and the discriminator learns the augmented images are part of the real image distribution that should be adapted for generating synthetic images \cite{zhao2020image}. Hence the generator learns to produce images with undesired augmentation artifacts such as noise, colour, cutout, and geometric operations. This learning practice and producing images with undesired augmentation artifacts are called \textit{“leaky augmentations”}.

A wide range of augmentations may be used to stop the discriminator from overfitting while ensuring that applied augmentations do not leak into the resulting generated images. In addition, an Adaptive control procedure may enable the model to function effectively irrespective of the volume of training data, the dataset's nature/characteristics, and the training approach.

\begin{figure}[htbp]
\centerline{\includegraphics[width=0.75\textwidth, angle=0 ]{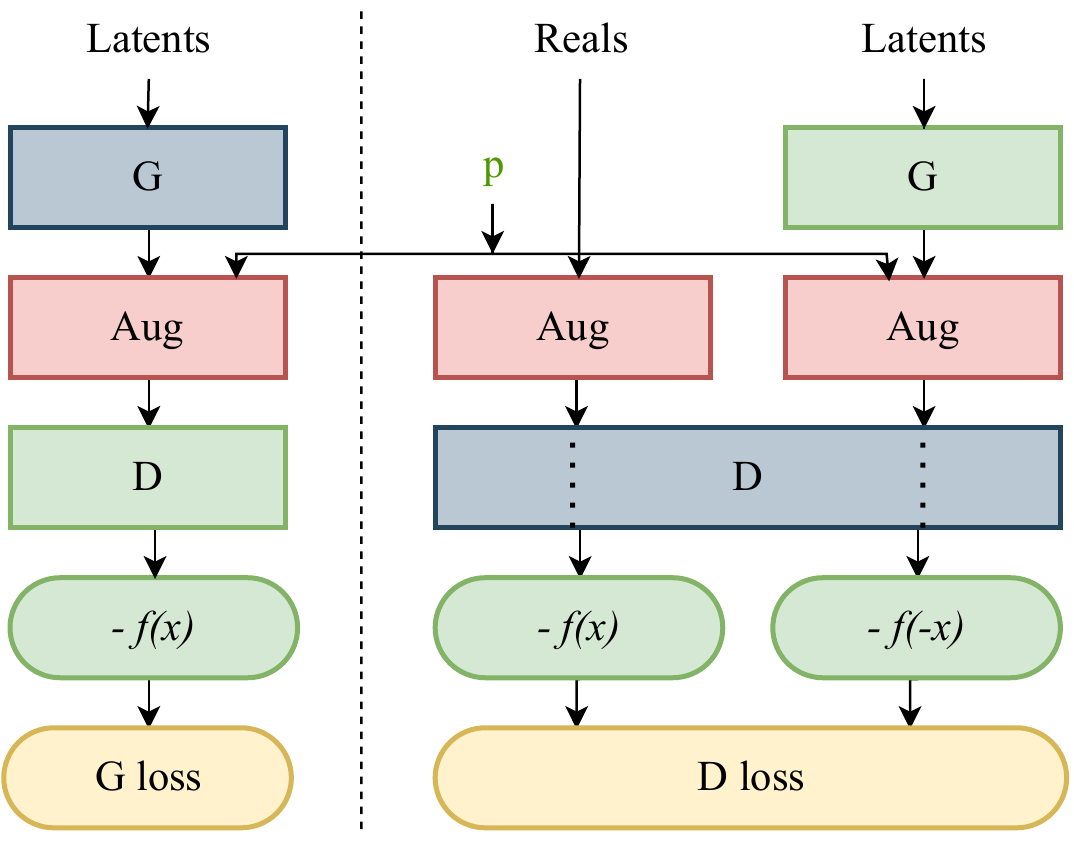}}
\caption{The flow diagram of Stochastic Discriminator Augmentation \cite{karras2020training}, where $G$ is the generator and $D$ is the discriminator. The red boxes represent the 18 augmentation operations. The set of selected augmentation are controlled by the augmentation probability \textit{'p'} and these augmentation can be visible to the discriminator $D$ in the green box. The blue boxes represent the networks that get trained during the training process and yellow boxes represent the loss calculated after the training. In this set up, the non-saturating logistic loss is accommodated to calculate the final probability of the images being predicted as fake. 
}
\label{fig3}
\end{figure}

Overfitting in various GAN models, especially in the variants of StyleGANs, can be observed when the value of the Fréchet Inception Distance (FID)\cite{heusel2017gans} metric starts to escalate without any decline, leading to leakage in the augmentations. To prevent such behaviour, a pipeline known as \textit{“Stochastic Discriminator Augmentation”} is introduced. This approach is inspired by the balanced consistency regularization (bCR) approach by Zhao et al.\cite{zhao2021improved}, designed to prevent leaking of the augmentations. Stochastic Discriminator Augmentation is a flexible type of augmentation that prevents the discriminator from becoming overly confident by showing all the applied augmentation to the discriminator. The discriminator is evaluated based on the augmented images, using the same augmentation as was applied when training the generator. In this practice, the discriminator can see the training images, which assists the generator in generating desired ideal outcome. Fig. \ref{fig3} shows the workflow of Stochastic Discriminator Augmentation.

Similarly, in order to regulate the distribution in the generated images, the idea of \textit{invertible transformation} is used. Invertible transformations are beneficial when applying a wide range of augmentations; for example, 18 types (clustered into 6 categories) of augmentations are used. Invertible transformation in the augmentation can be defined as, “for a target distribution $y$ and an augmentation operator $T$, the generated distribution $x$ is trained such that  the augmented distributions match with the target distribution $y$” \cite{karras2020training}. If a transformation is non-invertible there will be leakage but if all the transformations are invertible there will be no leakage. Invertible transformations can be reversed by the generator and removed from the distribution while non-invertible ones cannot be removed and can result in leakage. The generator network learns to generate the images in the correct underlying distribution by undoing the augmentation that does not fit the right kind of distribution. Hence, applying this concept of invertible transformation in augmentation \cite{bora2018ambientgan} helps with finding the correct target distribution of the data.

Another trick used to prevent leaking is to apply different augmentations in \textit{a particular fixed order}; for example, blitting, geometry and colour. Therefore, a sequential composition of augmentations that do not leak will ensure no leakage to the generated images. 

Although the Invertible Transformation process prevents the augmentation from leaking at least at the very early stages of training, which is desirable, a few constraints still require to be addressed. Augmentation leaking is highly dependent on a probability value, \textit{‘p’}. Higher values of \textit{‘p’} may confuse the generator by picking one of the random possibilities of the augmentation and image distribution; this phenomenon makes the chosen augmentations leak. If \textit{‘p’} is under a safety limit, it is less likely to produce leaking augmentation on generated images. To keep track of the safety limit value, an adaptive approach was introduced.

The concept of \textit{Adaptive Discriminator Augmentation} was supported by \textit{controlling the augmentation strength ‘p’} by which the augmentation is applied as the training progresses. The initial value of \textit{‘p’} starts from 0 and gets regulated in every 4 mini-batches as training progresses. If overfitting occurs during the training, the p-value can be adjusted by a fixed rate. A given target value can control the strength of the p-value. This concept of setting a target value, aka \textit{“ADA target”}, came from observing the training process and the safety limit of value \textit{‘p’}. For example, In the study by Karras et al. \cite{karras2020training} it was observed that the FID value declined after \textit{‘p’} became close to 0.5. Hence the ADA target was set as 0.6. Regardless of the dataset volume, discriminator overfitting was avoided by implementing this strategy, and convergence was achieved during the training.

Despite the fact that GAN models have been very sensitive towards hyperparameter selection, StyleGAN2-ADA supports reasonable quality of results without major changes in the hyperparameters and loss functions while training from scratch or performing transfer learning.

\subsection{The impact of $R_1$ Regularization \texorpdfstring{$'\gamma'$}{Lg} for 300 images}
\label{sec:The Impact of R1}
As discussed in section \ref{sec:StyleGAN2 with Adaptive Discriminator Augmentation}., one of the limitations of GANs is that small data may lead to overfitting, divergence or mode collapse. These grounds motivate our work to adapt StyleGAN2-ADA, which uses a minimum of 1000 images for experimental purposes. In this work, we used a limited amount of input images i.e. 300 images, but with fine-grained vital features i.e. Rosacea condition. Given the limited number of images, it might be hard to retain the most important features while training the networks and generating synthetic images. Hence, it is necessary to explore the strategies which may help obtain better results along with the adaptation of StyleGAN2-ADA. 

The StyleGAN2-ADA architecture functions very well even without changing network architectures, loss functions or other key parameters. As GANs are sensitive to hyperparameters, in this work most of the hyperparameters are kept unchanged except for the \textit{$R_1$ Regularization weight/strength} $'\gamma'$. According to a few studies, regularization has a significant impact on stabilizing GAN training. Regularization helps produce high-quality images by stabilizing the broad range of  noise levels \cite{roth2017stabilizing}. In the instances of images with a high number of features, $R_1$ Regularization (aka $L_1$ norm Regularization) performs satisfactorily in feature selection by removing some unimportant features. It helps in shrinking the coefficient of the less important features to 0. $R_1$ Regularization helps to prevent overfitting. To prevent overfitting due to the small volume of data, regularization extensively reduces the variance of the model without losing important attributes in the input image features and without a significant rise of bias in the model. On the contrary, after a certain numerical value for the strength $'\gamma'$, the model cannot capture the input images’ vital attributes. In this work, those particular numerical values of $'\gamma'$ are explored, with the aim of retaining vital details of the input images.

In GANs, Generator $G$ and Discriminator $D$ are the two modules/networks optimized by playing a Minimax zero-sum game with each other while the task is to learn the distribution of data. The distribution of images means the distribution of pixel values in a particular pattern, that makes all the images to align similarly. The task of the generator is to generate synthetic images that follow the same distribution as the input images and look as similar to the input images such that it is hard to differentiate between synthetic and real images. The task of the Discriminator is to differentiate between the real input images and the synthetic images created by the Generator. Hence the key goal of the Generator is to create images in such a way that the discriminator is deceived in finding out the difference between real and synthetic images. These events are regulated by a cost/value function, which is optimized during the training process. Hence the output of the discriminator is a cost function given by the negative log-likelihood of the binary discrimination task between real and synthetic images and another output is a probability of the images being real and synthetic. So the discriminator attempts to minimize this binary discrimination error, while the generator attempts to maximize this error. The binary discriminator error is directly proportional to the network ($G$ and $D$) loss. As the Discriminator error minimizes, the loss ($D$ loss) maximises; as the Generator error maximizes, the loss ($G$ loss) minimizes and this is the main goal of GANs. The equation below represents the cost/value function $V$ which the GANs optimise during training. In the equation, the first term only applies to real data and the second term only applies to synthetic data. $x_{real}$ represents a real image and $z$ represents the random input values/noise for $G$. The cost/value function $V (G, D)$ is defined as:

\begin{equation}
\begin{aligned}
    \min\limits_{G} \max\limits_{D} V(D,G) = \E_{x\sim p_{data}(x)} [\log(D(x))] + 
    \E_{z\sim p_{z}{(z)}} [\log(1-D(G(z)))]
    \label{eq1}
\end{aligned}
\end{equation}

As the GAN concept is based on the \textit{Zero-sum game}, it is expected to attain a \textit{Nash Equilibrium} in which each player cannot reduce their cost function without changing the parameters of the other  player\cite{fedus2017many}. As defined, equilibrium is a situation in which no player could improve its position by choosing an alternative available strategy(\textit{‘cost function’} in this case), without implying that each player’s privately held best choice will lead to a collectively optimal result \cite{holt2004nash}. 

The \textit{cost/loss/value function} is affected by the integrated $R_1$ Regularization. It is necessary to achieve the lowest divergence between the training distribution and the model distribution that obtains minimum loss at equilibrium. Despite this, it is hard to reach the closest point towards the equilibrium when the input images are short in supply. Hence it is essential to leverage the advantage of $R_1$ regularization strength to achieve minimum loss. The study by Mescheder et al.\cite{mescheder2018training} stated that using $R_1$ regularization helps in stable training as well as high-resolution image distribution for CelebA and LSUN datasets. Under suitable assumptions the $R_1$ regularization strength has an impact on obtaining notably better results in the generated image quality.

Hence this work examines the effects of  $R_1$ regularization to find the most favourable strength $'\gamma'$ that suits the nature of our dataset, since choosing the value of $'\gamma'$ is highly dependent on the dataset size and nature. However, a few studies have proposed a mathematical formulation to initiate the value of $'\gamma'$ as an initial guess in Equation \ref{eq2}; Where $N=w\times h$ (in this case $512\times512$) and $M$ is the size of minibatch, $w$ and $h$ are the number of pixels\cite{karras2020training, roth2017stabilizing}. In the study by Mescheder et al.\cite{mescheder2017numerics} on the impact of regularization, even though only a handful of images were used, the authors proved that an appropriate choice of $\gamma$ leads to better convergence properties near local Nash-equilibrium, which further leads to the generation of high-fidelity images while preserving fine-grained details learned from the input images. 

\begin{equation}
\begin{aligned}
    \gamma_0 = 0.0002 \cdot N/M 
    \label{eq2}
\end{aligned}
\end{equation}

\subsection{Rosacea datasets}
\label{sec:Rosacea datasets}
GANs have produced impressive results due to the availability of the enormous volume of images on various web sources, which have relaxed terms of privacy and copyright. Most of the large datasets used in the improvement and study of GANs contain objects, animals, paintings or faces of celebrities. StyleGAN2-ADA uses histopathological images of breast cancer, which do not disclose patients' identities. Similarly, some other imaging modalities such as dermoscopic imaging, X-Ray imaging, and MRI scans may not disclose the person's identity. Especially when a skin condition is captured directly, focusing on the affected region of the body, a person's identity is hardly identifiable. However, in the case of full facial images with skin conditions such as Rosacea, capturing the entire face can result in identifying the patient. 

\subsubsection{Publicly available data}
\label{sec:Publicly available data}
A few teledermatology web sources support computer-aided skin disease diagnosis research and development. The available Rosacea images in various web sources are listed in our previous work \cite{mohanty2022skin}. There are about 208 Rosacea images in total. Among these there are only a few images with full-face visibility and a few others are watermarked, which may affect the features in the generated images. In order to examine the nature of Rosacea in the facial region, it is essential to access high-quality full face images which are rarely found in online teledermatology sources. Hence, acquiring full-face images of Rosacea is a difficult task. 

\subsubsection{Rosacea Dataset- ‘rff-300’}
\label{sec:Rosacea Dataset- ‘rff-300’}
Acquiring a large volume of medical/ clinical images of facial skin conditions, including Rosacea, may be a time-consuming task. Moreover, there are privacy concerns to be addressed while distributing such images. Hence data acquisition is the main obstacle in this research. In this study, we have access to a small dataset, which is referred to as the “Irish Dataset” in the rest of the paper. The \textit{“Irish Dataset”} is provided by \textit{The Powell Lab, University College Dublin}\cite{powellLab}. The dataset contains 70 high-quality full-face images of Rosacea. The original images were present in various resolutions ranging from $800\times1000$ to $900\times1200$. These were later resized for the experiments. Among the 70 images in the Irish Dataset, 67 images were selected for experiments conducted in this study. 

Given the low number of images in the Irish dataset, it was essential to collect more data from various web sources, i.e. the teledermatology web sources and other Google search results. Thus, another 67 full-face images have been taken from SD-260\cite{sun2016benchmark}. A few more images were obtained from Google search results and teledermatology websites in accordance with the following criteria/standards: 
\begin{itemize}
    \item The resolution is a minimum of $250\times250$.
    \item visibility of full face including forehead to chin and both cheeks. 
    \item The images are labelled/captioned/described under subtypes 1 and 2. 
\end{itemize}

Given these standards/criteria, many Rosacea labelled images with partially visible faces are not considered in this study. This results in a total number of 233 full-face images with Rosacea acquired from publicly available sources. Finally, combining this with 67 images from the \textit{“Irish Dataset”}, provides us with 300 images for the experiments to generate synthetic full-face images with Rosacea. These 300 images are full front-view facial images with Rosacea subtype 1 or subtype 2. 

All 300 images are centre-cropped manually while preserving the visibility of the face and eliminating unnecessary background details and accessories around the ears and heads. The images are resized to $512\times512$ pixels to keep the optimum details of the disease. The preferred file format type “.png” was chosen to preserve the best possible sharpness of the original images. For ease of understanding and usage, the entire dataset used in the experiments is referred to as \textbf{“rff-300 (Rosacea-full-face-300)”}. 

\subsubsection{Implementation Specifications} 
\label{sec:Implementation Specifications}
A system equipped with an Nvidia Geforce RTX 3090 (24GB) GPU, an AMD Ryzen 9 5900X 12 core CPU, and 32 GB RAM are used to carry out the experiments. The complete implementation was carried out on Pytorch 1.7.1. with CUDA version 11.1 on Linux.

\section{Experiments and Results}
\label{sec:Experiments and Results}

The implementation choices in this work are the same as in the original work on StyleGAN2-ADA with some minor changes in the configuration. As the original work claims to have chosen the ideal configuration in network architecture and loss functions, hence these units are kept unaltered in these experimental implementations. The learning rate of 0.0025 is kept unchanged to examine the effect of augmentation and other existing hyperparameters on the output. All the 300 input images with resolution $512\times512$ are x-flipped, which brings the number of input images to 600. 

In most cases, the augmentation choices are limited to pixel-blitting and geometric augmentation, because other augmentations such as colour, filter, noise and cutout may affect the desired features of the disease. For instance, in the Transfer Learning set-up, the augmentations were applied too quickly at the early stages of the training. At the very beginning stage of the implementations and setup, a few experiments were carried out with all the given augmentations offered by StyleGAN2-ADA. However, a set of augmentations such as colour, filters, noise and cutout started to leak at the later stages of the training. One of the augmentations which had shown frequent leaking was the colour augmentation. This problem was also encountered in the work by Karras et al.\cite{karras2020training}. Hence those experiments and results are not included in this study. The further experiments were set up with a limited set of augmentations and those experiments are listed in Table \ref{tab1}. 

As in this work a 24 GB GPU was used for the experiments, and several configuration choices required adjustment and recalculation during the experiments. The minibatch size, mini-batch standard deviation, exponential moving average, $R_1$ regularization $\gamma$ were altered according to the nature of the input and GPU configuration. The alterations on these hyperparameters are dependent on image resolution and GPU model. The numeric value of these hyperparameters helps in reducing computational space, time, and cost by leading to smoother progress during the training. 
As the input images resolution is $512\times512$ and the number of GPUs used is 1, the following configurations were used during the training
\begin{itemize}
    \item the minibatch size = max (min (1 $\cdot$ min (4096 // 512, 32), 64), 1) = 8,
    \item mini-batch standard deviation = min (minibatch size // GPUs, 4) = 4, 
    \item Exponential Moving Average = minibatch size $\cdot$ 10 / 32= 2.5
\end{itemize}
Among the various implementation choices, $R_1$ Regularization weight was given utmost importance during the experiments, which will be discussed in further sections. 

It is important to measure the image generation quality of synthetic images. Although the majority of experiments in StyleGAN2-ADA\cite{karras2020training} in the literature have been evaluated using FID; in this study, the experimental results were assessed using Kernel Inception Distance (KID)\cite{binkowski2018demystifying}. Lower values of KID indicate better performance. The main reasons to consider KID for the experiments are listed below: 
\begin{itemize}
    \item KID functions outperform FID in case of limited samples i.e., a small number of images. \item KID has a simple, unbiased, and asymptotically normal estimator, in contrast to FID. 
    \item KID compares skewness as well as mean and variance. 
\end{itemize}

\begin{table*}[!t]
\caption{\label{tab1}List of experiments and results}
\centering
\begin{tabular}{|p{1.00cm}|p{2.20cm}|p{1.5cm}|l|p{1.00cm}|p{1.50cm}|p{1.00cm}|}
\hline
Exp \newline no. & Training \newline set-up & Freeze-D & Augmentation Choice &
$\gamma$ & Best KID$\times 10^{3}$ \newline achieved & At step no.\\
\hline
1 & From scratch & NA & blitting, geometry,
\break colour, filter,
\newline noise, cutout & 6.5 & 6.8 & 2640\\
\hline
2 & From scratch & NA & blitting, geometry & 10 & 11.8 & 720\\
\hline
3 & Transfer \newline Learning (TL) from FFHQ & NA & blitting, geometry & 6.5 & 3.6 & 120\\
\hline 
4 & TL from FFHQ & 4 & blitting, geometry & 6.5 & 3.5 & 80\\
\hline 
5 & TL from FFHQ & 13 & blitting, geometry & 6.5 & 3.3 & 680\\
\hline 
6 & TL from FFHQ & 13 & blitting, geometry & 10 & 104.6 & 840\\
\hline 
7 & TL from FFHQ & 13 & blitting, geometry & 3 & 3.1 & 80\\
\hline
8 & TL from FFHQ & 13 & blitting, geometry & 2 & 4.2 & 360\\
\hline
9 & TL from FFHQ & 17 & blitting, geometry & 6.5 & 3.3 & 800\\
\hline 
10 & TL from FFHQ & 10 & blitting, geometry & 6.5 & 2.5 & 160\\
\hline
\multicolumn{7}{@{}l}{}
\end{tabular}
\end{table*}

\begin{figure}[htbp]
\centerline{\includegraphics[width=0.90\textwidth, angle=0 ]{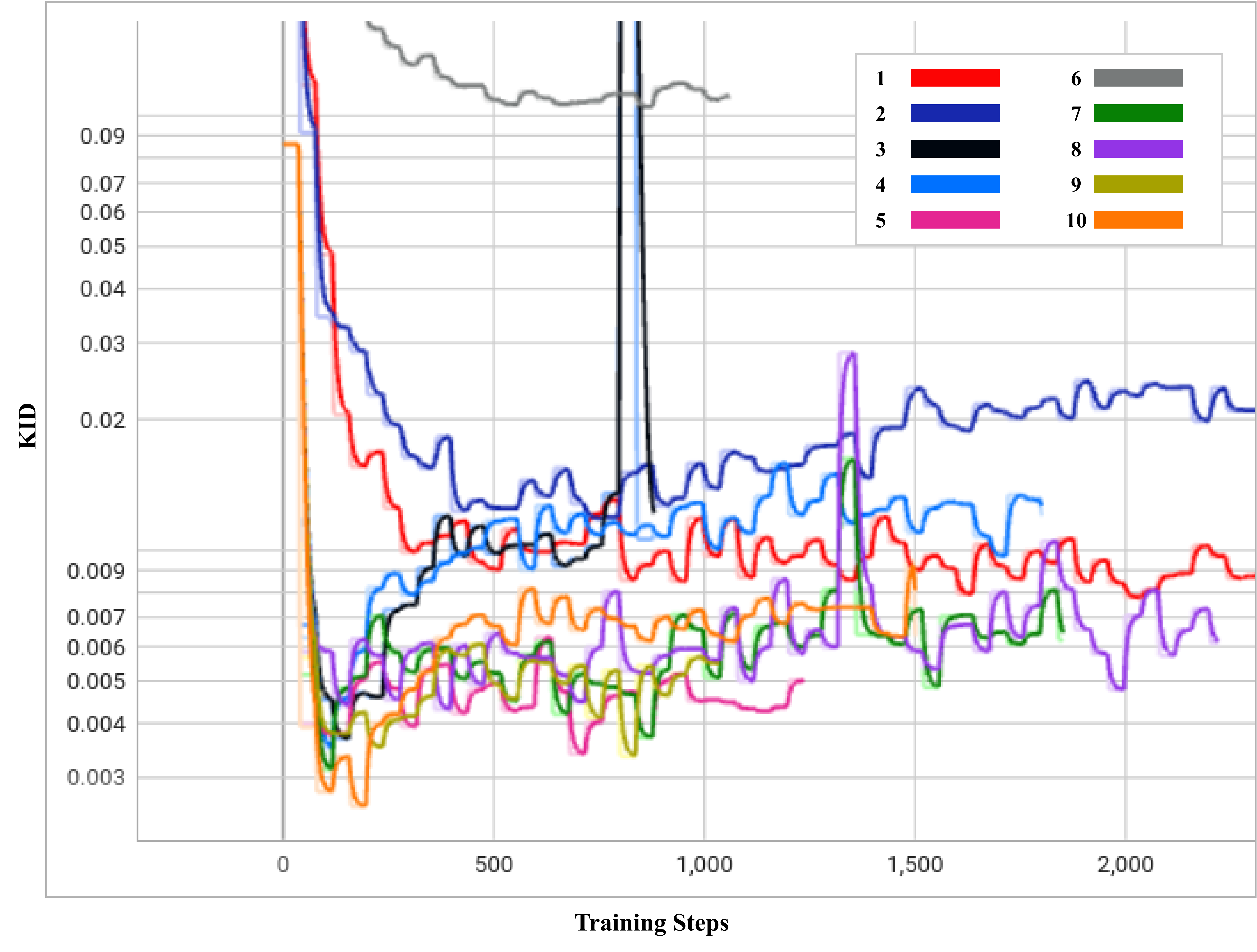}}
\caption{Progress of KID for 10 experiments over the training periods outlined in Table \ref{tab1}}
\label{fig4}
\end{figure}

\begin{figure*}[t!]
\centering
\includegraphics[width=0.90\textwidth, angle=0]{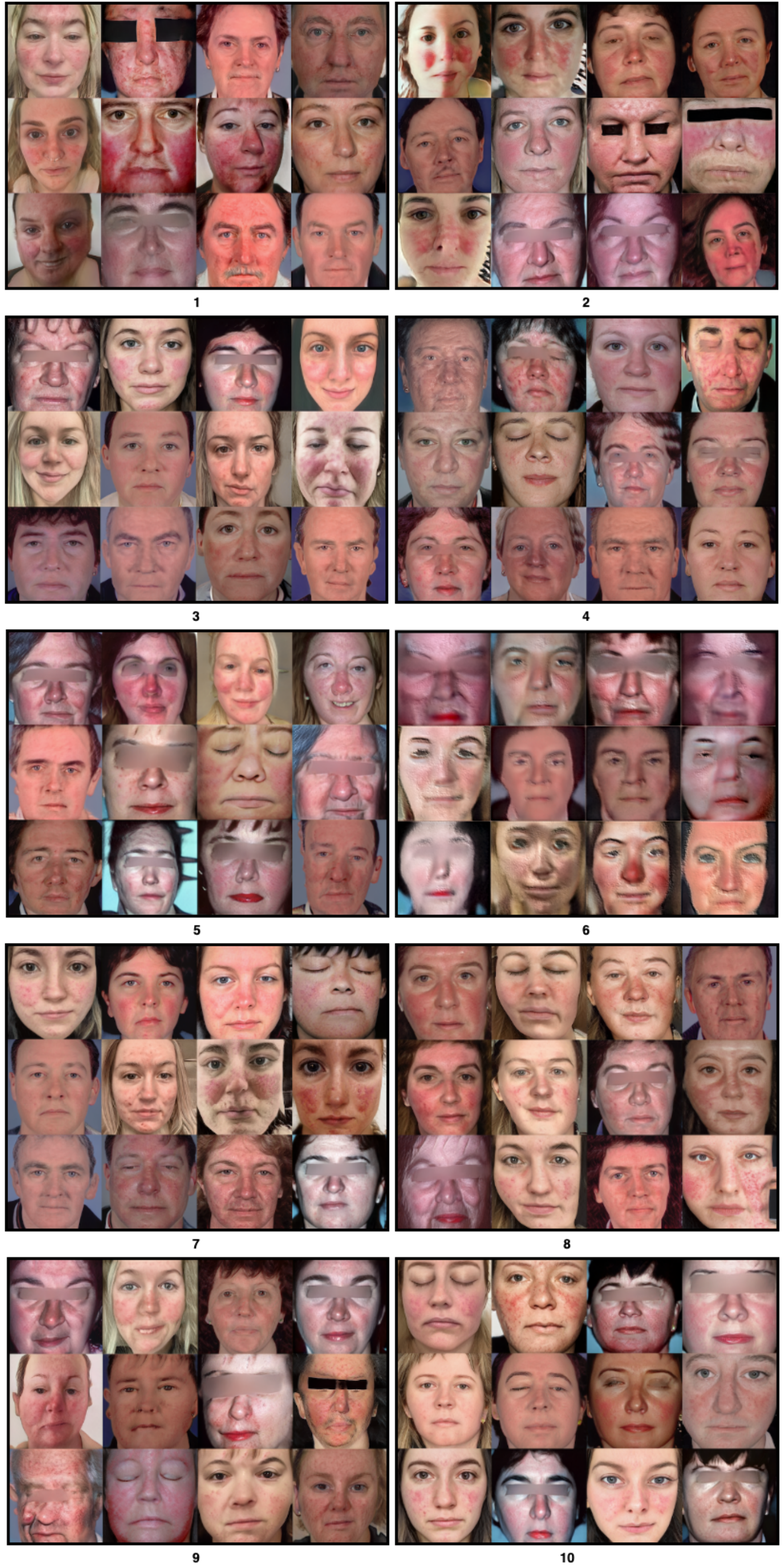}
\caption{Generated faces from 10 experiments outlined in Table \ref{tab1}}
\label{fig5}
\end{figure*}

As listed in Table 1, there are various experimental set-ups explored to obtain high-quality synthetic faces with Rosacea. The rationale for chosen parameter values and main findings are outlined below:
\begin{itemize}
    \item Training from scratch in \textbf{Exps 1} and \textbf{2} does not provide any advantage with the limited data i.e., 300 input images. However, these experiments show that the $\gamma$ value has a significant impact in terms of image generation and convergence during the training. As shown in Fig. \ref{fig4}, Exp 1 achieved the lowest KID at training step 2640 with $\gamma$ =6.5, whilst Exp 2 achieved the lowest KID at training step 720 with $\gamma$ =10. The distribution of Rosacea artefacts on the generated images from Exp 1 are better compared to the generated images from Exp 2. Hence, it can be concluded that Exp 1 has the best achieved KID and better-quality generated images when training from scratch; conversely, Exp 2 converged faster but generated lower quality images. A lower strength of $\gamma$ performed better for training from scratch.
    
    \item In contrast, Transfer Learning from FFHQ\cite{karras2019style} in \textbf{Exp 3} performed approximately 33 times better with the improvement in training time/cost and nearly twice better at the training step 120 with the lowest recorded KID value during the training with the $\gamma$=6.5. As the FFHQ dataset is fundamentally a facial dataset, it was expected to have a wide range of facial features in the resulting generated images. The generated images have shown a great level of improvement, although image generation quality can be further improved by freezing the top layers of the discriminator to preserve the smaller features of the disease.
    
    \item In \textbf{Exp 4}, along with Transfer Learning from FFHQ dataset, the \textit{Freeze-Discriminator (Freeze-D)}\cite{mo2020freeze} technique was studied to improve the fine-grained details of Rosacea in the synthetic faces. In this experiment, the top 4 layers of the Discriminator were frozen, which improved the result faster, compared to the Transfer Learning without Freeze-D technique. The augmentation choice was kept unchanged to the previous experiment i.e. pixel blitting and geometric transformations. The $R_1$ regularization weight is set to 6.5. Figure \ref{fig4} represents the obtained KID values during the training process, in which the best value of KID = 3.5 is achieved at the step 80. Hence, it is observed that the training process improves relatively faster when the top layers of the discriminator are frozen. As Transfer Learning with Freeze-D presented better results, that offered motivation to explore various arrangements of Freeze-D. 
    
    \item Further the Freeze-D technique with Transfer Learning was applied by freezing 13, 10 and 17 layers of Discriminator. In \textbf{Exp 5}, the 13 top layers of Discriminator were frozen during the training with the same settings for augmentations i.e., pixel blitting, geometric transformation and $\gamma$ =6.5. The outcome of this experiment is inferior compared to the previous experiment, based on the inconsistency in training and the lowest KID achieved at the later stage of the training i.e. 3.3 is achieved at training step 680. The generated images from this experiment were lower in quality, e.g. most of the facial features are deformed, blurred with leaky background details. To improve this condition, further experiments were carried out with higher and lower strengths of $\gamma$ while keeping other hyperparameters unchanged. 

    \item Although some higher values of $\gamma$ were tested while training from scratch in exp 2, they were not used with Transfer Learning, hence $\gamma$ =10 was tested in \textbf{Exp 6}. It can be observed from Fig. \ref{fig4} and Table \ref{tab1}, that it took longer to achieve a minimum KID at step 840. The lowest obtained KID in this experiment was the highest KID value recorded among other experiments, proving it the worst KID value recorded. The generated images in Fig. \ref{fig5}(6) were highly distorted and unusable in quality. However, it demonstrated the significance of $R_1$ regularization strength $\gamma$. Regardless of training set up, higher values of $\gamma$ performed worse in terms of convergence and quality of generated images. 
    
    \item Hence in the next experiments, the lower values of $\gamma$ were explored. In \textbf{Exp 7} $\gamma$ =3 was examined, while other hyperparameters were kept unchanged as in the previous Exp 6. As was observed in Fig. \ref{fig4}, KID drops at the very beginning stage of training i.e. step 80 and then becomes inconsistent. However, this is one of the second lowest KID values achieved among all the experiments resulting in high quality images generated at step 80 with the KID value 3.1. The generated images were with fine-grained details of Rosacea and disease patterns and resembled the real-life cases of Rosacea.
    
    \item To exploit the performance with lower values of $\gamma$, \textbf{Exp 8} was carried out with $\gamma$ =2. In this experiment, the lowest KID = 4.2 was recorded at training step 360. It was observed from Fig.\ref{fig5} (8) that the generated samples are deformed at the left bottom portion with the blurred edges. The distribution of the disease feature was inadequate. It was observable that a low value of $\gamma$ produces such a strong sort of deformity which was not encountered in the previous experiments. 
    
    \item Furthermore, experiments \textbf{Exps 9} and \textbf{10} were carried out by freezing 17 and 10 layers respectively with $\gamma$ =6.5 to observe changes due to freezing the layer of Discriminator. Exp 9 shows inconsistency throughout the training process from the beginning. The minimum KID=3.3 is obtained at training step 800. In Fig. \ref{fig5}(9), it is observed that the generated images tend to be blurred around the edges and the center. Some samples are negatively affected by the geometric augmentation. 
    
    \item In \textbf{Exp 10}, generated sample images at the best value of KID =2.5 were obtained at the training step no.160. Although Exp 10 has obtained the lowest KID among all the experiments, the generated images are blurred at the edges and center as depicted in Fig.\ref{fig5}(10). The details of Rosacea are absent. 

    \item As the Freeze-D technique with freezing 4, 10, 13, 17 layers of Discriminator were experimented; the results showed that freezing 10 layers helps achieve the lowest value of KID amongst other training setups. However, it is observed that freezing 10 layers leads to too much smoothing which does not help in preserving the details of the disease. Freezing 4,13,17 layers of Discriminator achieved comparatively better results in terms of the value of KID. 
    
    \item Along with freezing the layers, we have experimented with various strength of $R_1$ regularization. Adapting various $\gamma$ values illustrates its significant impact on the training process, the metric (KID) and the generation of synthetic images.
    
    \item The impact of $\gamma$ value can be observed in both settings such as training from scratch and in Transfer learning. Exps 2 and 6 were carried out with higher strength of gamma and they have demonstrated the significance of the value very distinctly. The lower value of $\gamma$ leads to better results in training, given other implementation choices kept unchanged. 
    
    \item The choice of $R_1$ regularization weight/strength $\gamma$ value depends on the input data. There is a heuristic formula in \ref{eq2} for choosing the numerical value of $\gamma$ as an initial guess, which calculates the $\gamma$ value as 6.5. However, tweaking/adjusting this numerical value leads to better results in generating synthetic images with fine-grained details and improved fidelity. It can be acknowledged that the choice of $\gamma$ value is sensitive when the number of images are short in supply. Lower values of $\gamma$ perform better compared to the value obtained by applying the heuristic formulae. However, there is a risk in choosing very low values or very high values. 
\end{itemize}

%%%%%% INSERT THE FIGURE HERE

\subsection{Truncation Trick}
\label{sec:Truncation Trick}
The \textit{Truncation trick} was introduced by BigGAN \cite{brock2018large}. The truncation trick acts as a boosting strategy for the quality of images. By applying the truncation trick, we can expand the span in the variety of images. The quality of these individual images is comparatively high, and the distribution of disease artefacts is precise. Transforming the images to latent space provides an opportunity to generate 1000 high-quality synthetic images at a time. This is possible with the Truncation trick introduced by BigGAN architecture. 

The truncation trick is a sampling technique that aims at truncating the noise vector $z$ by resampling the values to improve individual sample quality. The truncation trick is regulated by a value called the \textit{‘truncation threshold’} $\psi$. The truncation threshold can lie in the range between 0.5 to 1. As per \cite{karras2020analyzing, brock2018large}, we have used a truncation threshold $\psi$ of 0.7 to obtain the optimal results. Choosing the truncation value of 1 indicates that there is no truncation. Different truncation thresholds help in truncating the latent values so that they fall close to the mean. The smaller the truncation threshold, the better the samples will appear in terms of variety.  

Although \textbf{Exp 10} has achieved the lowest value of KID, the images generated from this experiment are \textbf{not useful} due to a few factors made, such as: 
\begin{itemize}
    \item A few images were not properly distributed and they are distorted and blurred with leaked geometric augmentations, 
    \item While exploiting the latent space, most of the samples generated from this experiment lacked variation in regards to common facial features as well as Rosacea features,
    \item As a result, out of 1000 generated images, only 30 high quality images were picked for further analysis.
\end{itemize}

On the other hand, \textbf{Exp 7} achieved the second lowest value of KID, the generated images from this experiment were \textbf{useful} due for a few reasons, such as: 
\begin{itemize}
    \item All 1000 sample generated (from step 80 with the best KID) were correctly distributed,
    \item The span of variation was greater than Exp 10, meaning that there was more variety in facial features and Rosacea features, 
    \item There were no deformations in the facial and Rosacea disease features,
    \item The samples were not highly smooth in the forehead or cheeks region, 
    \item More distinctive facial and Rosacea disease features obtained compared to Exp 10,
    \item As a result, the best 300 high quality images were picked through visual scrutiny from Exp 7.
\end{itemize}

Figure \ref{fig6} and Figure \ref{fig7} are the generated images through truncation from Exps 7 and 10 respectively. These 300 synthetic images selected from Exp 7 were used for further qualitative analysis discussed in Section \ref{sec:Qualitative Evaluation}. These 300 images are named as synthetic rosacea full faces \textbf{(synth-rff-300)} are available on: \url{https://github.com/thinkercache/synth-rff-300}

\begin{figure}[htbp]
\centerline{\includegraphics[width=0.80\textwidth, angle=0 ]{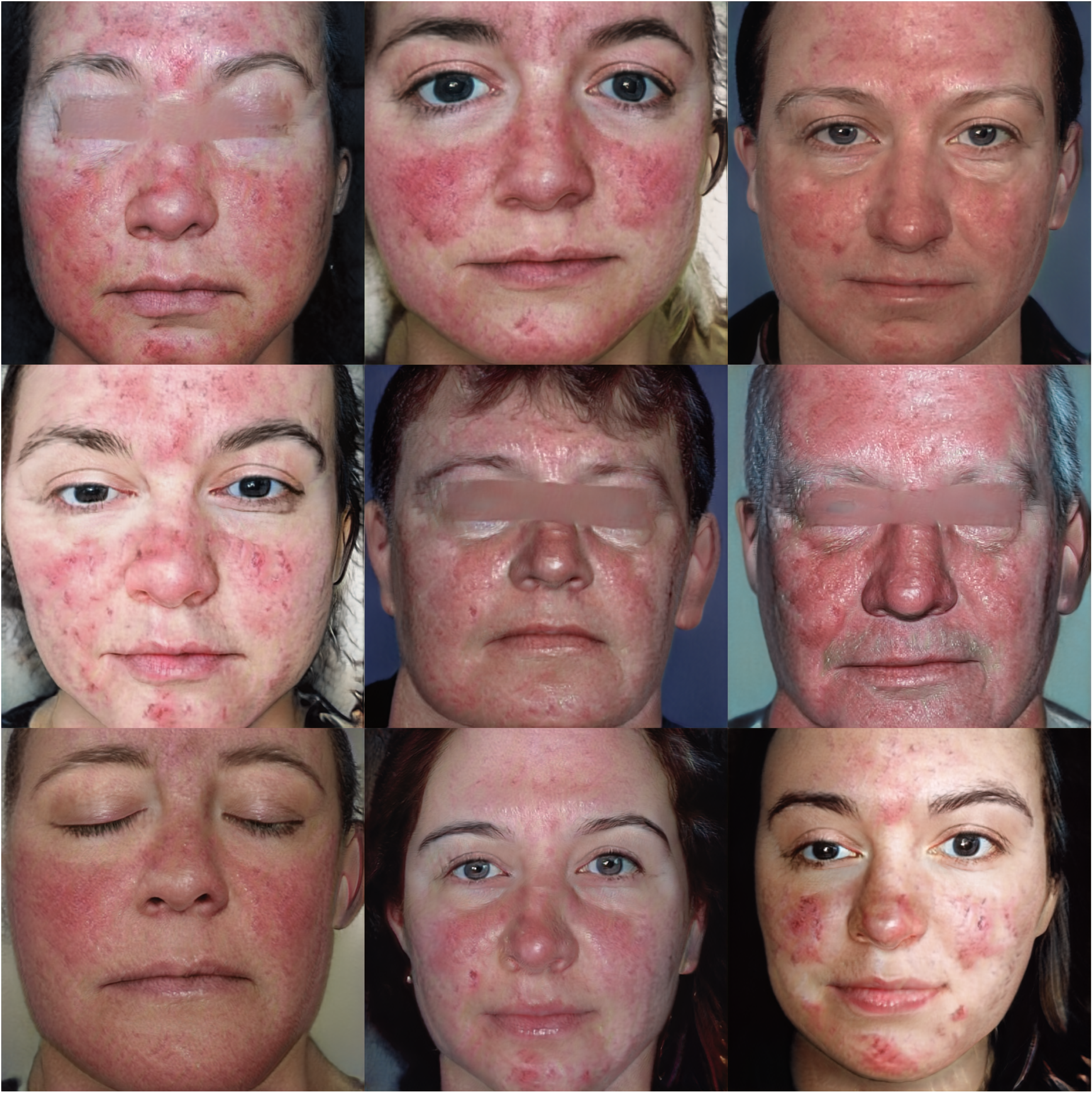}}
\caption{Generated faces from the best KID value (3.1) of Exp 7 with the truncation $\psi$=0.7 }
\label{fig6}
\end{figure}

\begin{figure}[htbp]
\centerline{\includegraphics[width=0.80\textwidth, angle=0 ]{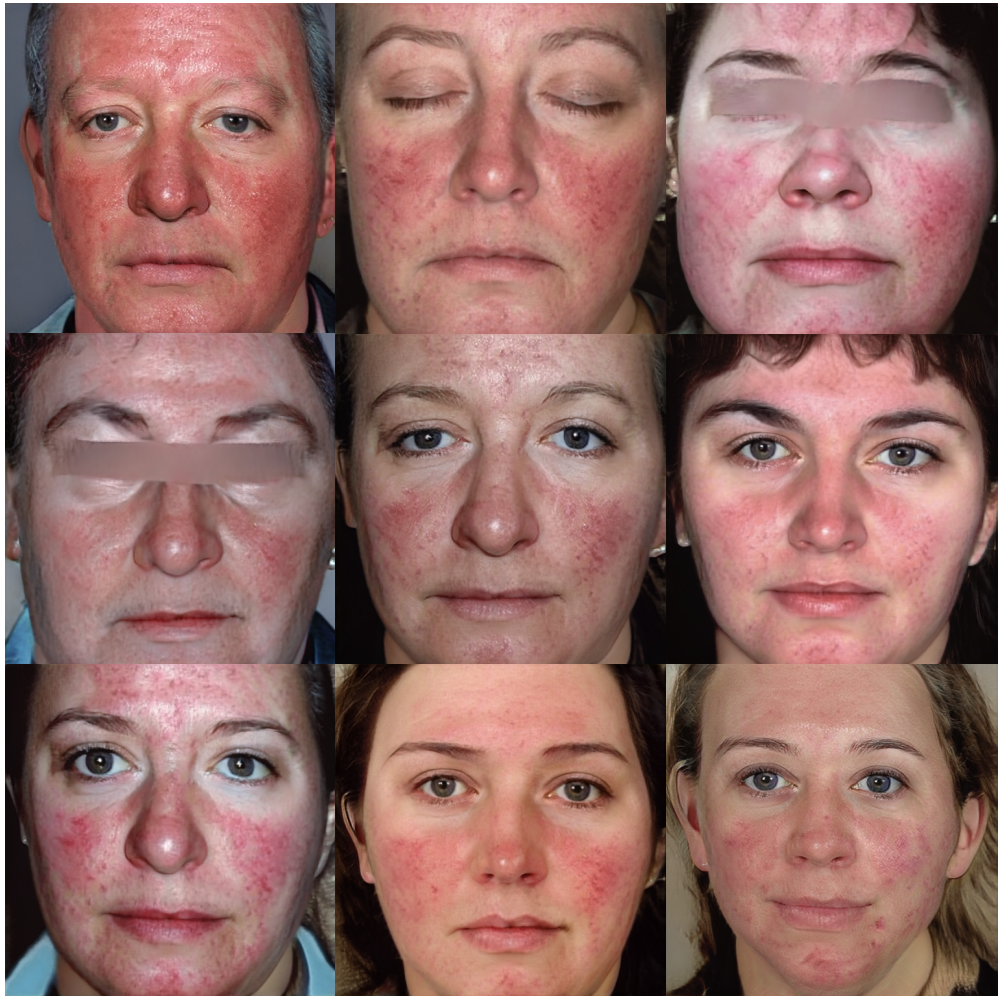}}
\caption{Generated faces from the best KID value (2.5) of exp 10 with the truncation $\psi$=0.7}
\label{fig7}
\end{figure}

\section{Qualitative Evaluation of generated images by the specialist dermatologist and non-specialist participants}
\label{sec:Qualitative Evaluation}
\begin{figure}[htbp]
\centerline{\includegraphics[width=0.75\textwidth, angle=0 ]{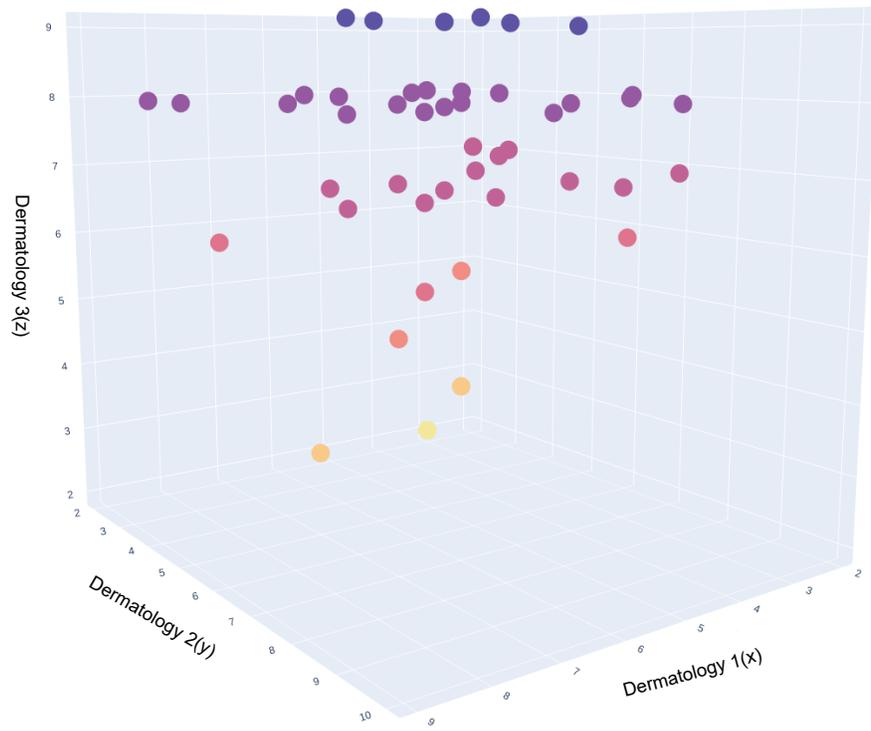}}
\caption{A 3D representation of Dermatologists opinion on synthetic images}
\label{fig8}
\end{figure}

Although the best 300 high quality images with resolution $512\times512$ were selected from the Exp 7, it is important to get them verified by dermatologists to validate the feature and distribution (location/colour/nature) of the Rosacea. However, inspecting all the 300 synthetic images is a time-consuming task. Hence out of 300 images, about 50 images were randomly picked for the inspection by the \textbf{expert dermatologists}. The images were organised in a Google form. The dermatologists were requested to rate the images from a medical perspective as to how well the artefacts on the generated faces represented Rosacea on a linear scale from 1 (not realistic Rosacea) to 10 (very realistic Rosacea). In total, three dermatologists participated in this study. The scatter plot in Fig. \ref{fig8} illustrates the average rating over the three dermatologists per image. The dots in this 3D plot represent the synthetic images. The darkest colours represent the images with higher ratings followed by the lighter shades for the lower ratings.

\begin{figure}[htbp]
\centerline{\includegraphics[width=0.80\textwidth, angle=0 ]{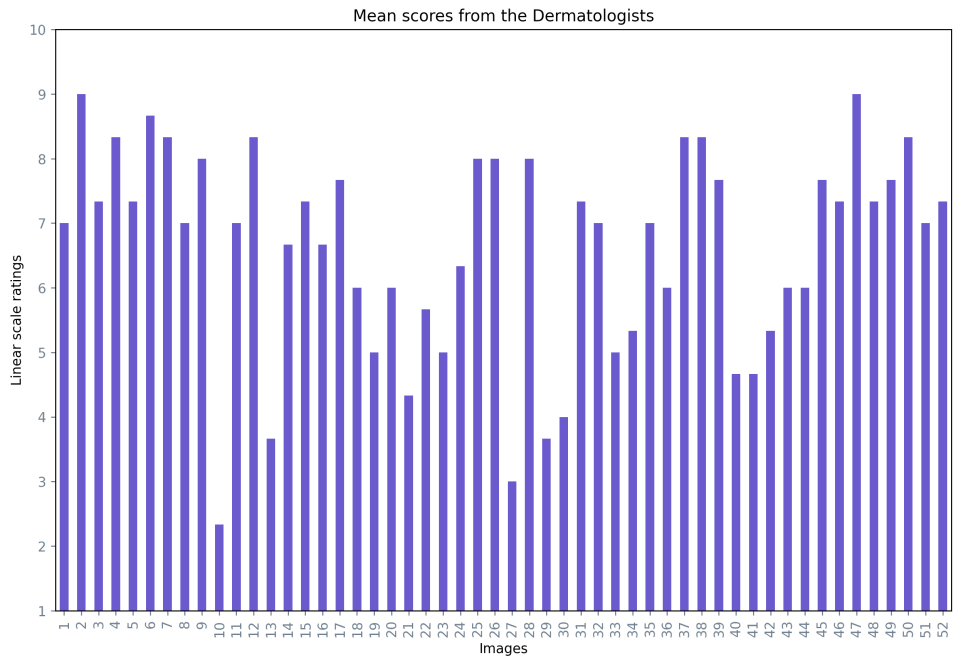}}
\caption{Mean scores from the Dermatologists for generated images}
\label{fig9}
\end{figure}

Fig. \ref{fig9}. presents the mean score for each image averaged over the three dermatologists. 73\% of the images had a mean score of over 60\%.  

\begin{figure}[htbp]
\centerline{\includegraphics[width=0.80\textwidth, angle=0 ]{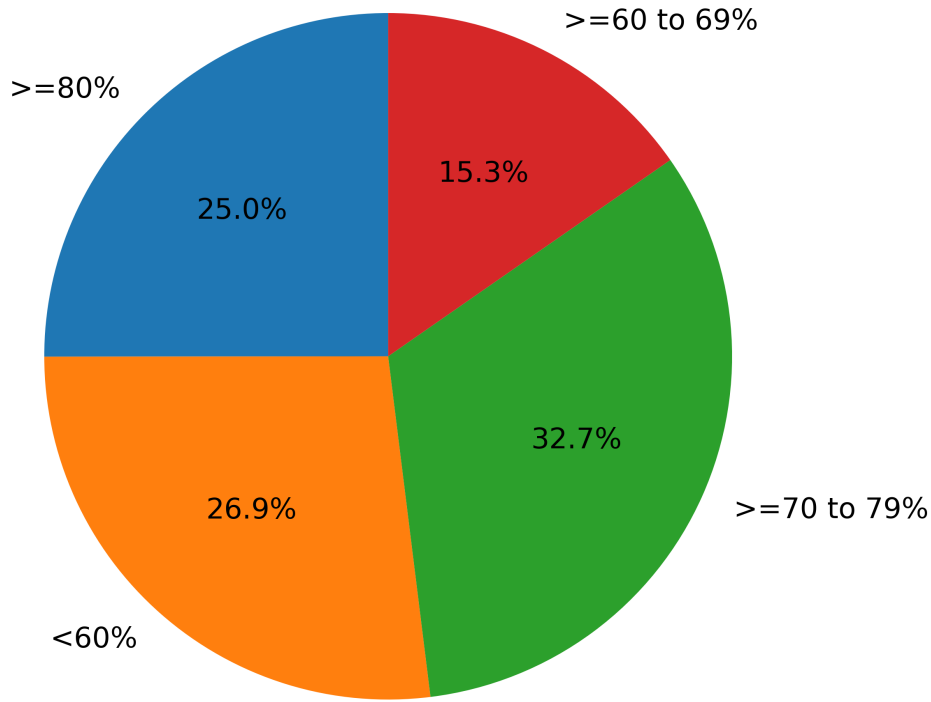}}
\caption{Representation of mean scores (in \%) for percentage of given images in the study by Dermatologists}
\label{fig10}
\end{figure}

Out of 73.07\% of images (with more than 60\% mean score), 25\% of images were rated greater than 80\% mean score, 32.7\% images were rated greater than 70\% to 79\% mean score and 15.3\% were rated greater than 60\% to 69\% mean score values as depicted in Fig. \ref{fig10}.

To summarize, according to the dermatologists’ opinions (in medical perspective), 73\% of images present a realistic pattern of Rosacea on the generated faces and additional comments provided by the dermatologists are listed on the Table \ref{tab2}. Table \ref{tab2} \textbf{concludes} that the experts' overall impression of the generated Rosacea images is very positive. The remarks given by the experts ensure that developing synthetic images help in overcoming the data-scarcity problem for Rosacea and many other facial skin conditions in the medical imaging domain. 

The amalgamation of methodology for synthetic face generation, and from the quantitative and qualitative data, shows an optimistic direction for synthetic data generation for rare skin conditions and other diseases that involves medical imaging. This strategy will help deal with data scarcity problems in many disease domains and facilitate early and faster diagnosis.

\begin{table}
\centering
	\caption{Dermatologists comments on generated Rosacea faces}
	%\label{table}
	%\setlength{\tabcolsep}{3pt}
	\begin{tabular}{|p{55pt}|p{300pt}|}
		\hline
		Dermatologists & Comments \\ \hline
		1 & “Diagnosing Rosacea in some patients requires running a lab examination. But, essentially the images in this research created using an artificial intelligence can widely impact the performance of the technologies currently available to dermatologists. I believe these images could also be used for educational purposes if provided with a set of controls to create more variations of the disease. Best of luck.” \\ \hline
		2 & “I am surprised to see what AI can do. I think this work may help in Rosacea screening later on.” “A few images had a strange form of distortion on the face region but, in general, I am very surprised by the quality of the images and varying intensity of Rosacea in each image.” \\ \hline
		3 & “Please note, I have only examined the Rosacea and without taking notice of the other characteristics of the faces. I can say ETR is very realistic indeed. Great work, all the best.” \\ \hline
	\end{tabular}
	\label{tab2}
\end{table}

\begin{figure}[htbp]
\centerline{\includegraphics[width=0.95\textwidth, angle=0 ]{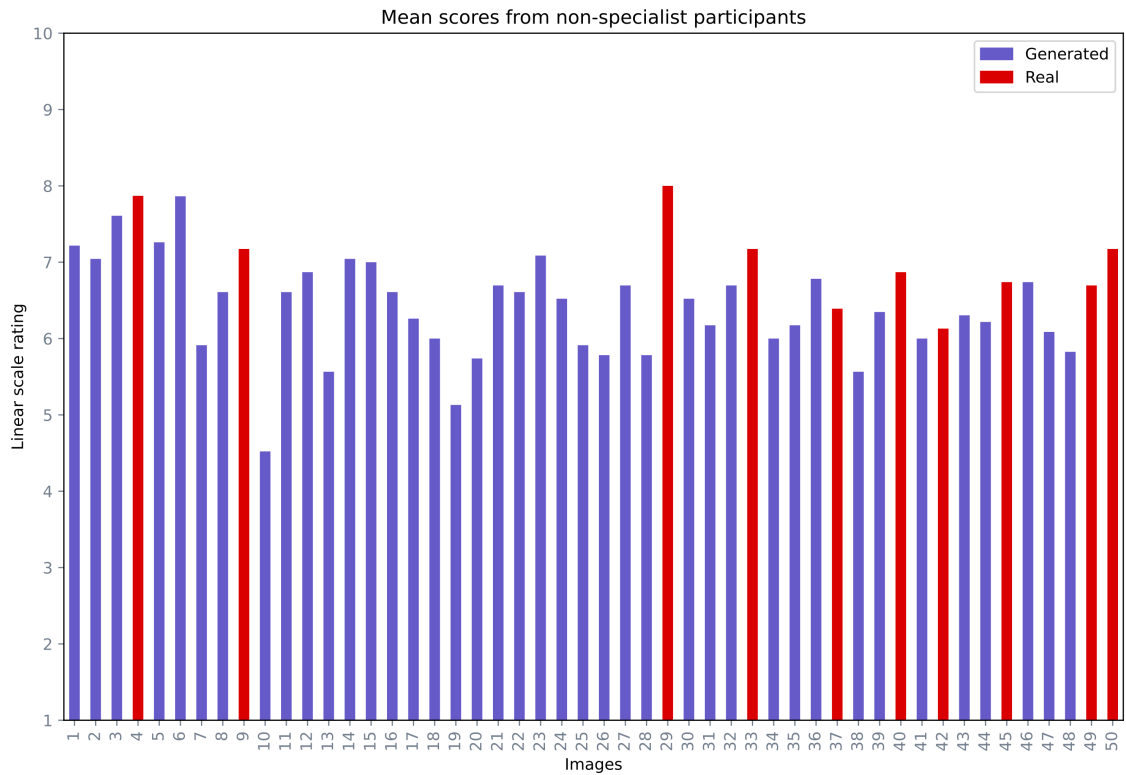}}
\caption{Mean scores from the non-specialist participants for miscellaneous images}
\label{fig11}
\end{figure}

The second part of the qualitative evaluation was based on  \textbf{non-specialist participants’} opinions. In this analysis, a total of 50 images were provided for analysis in which 40 images were generated and 10 images were real. The intention of including 10 real images was to analyse if non-specialist participants could see the difference between the real and fake images. The non-specialist participants were requested to rate the images in the range from 1 (not a realistic face) to 10 (a very realistic face). Fig. \ref{fig11} depicts the mean score range of each image, where generated and real images were labelled in different colours. Out of 50 images, 40 images got the mean score equal to and greater than 60\%. Among the top 10 images with highest mean score, 5 images (29, 4, 9, 33, 50) are real and 5 images (6, 3, 5, 1, 23) are generated. 

\section{Limitations and Discussion}
\label{sec:Limitations and Discussion}
Quantitative evaluation of generated images by GAN models particularly in medical imaging is an open-ended problems. Thus, various quantitative and qualitative methods have been adapted and are still in the development stage \cite{borji2022pros}. The quantitative evaluations are often performed using various metrics such as Inception Score (IS), Fréchet Inception Distance (FID), Kernel Inception Distance (KID), Precision-Recall, and Perceptual Path Length. These metrics are proven to function adequately with certain types of popular datasets which are large in quantity. Although such methods are designed to assess the quality of images or evaluate the distribution of the generated images, they may not be a reliable measure for applications in the field of dermatology. These metrics fail to provide any information regarding the "quality" of the generated artifacts on the skin which is vital in diagnosing skin conditions. In the field of dermatology, a minor change on the skin could be meaningful. The existing numerical methods are not capable of measuring the realism of the generated artifacts on the skin and whether they represent a skin condition or not.
 
As discussed in Section \ref{sec:Rosacea Dataset- ‘rff-300’}., this research utilises a limited dataset with 300 images to train a generative model. Following the state of the art studies, we deployed a quantitative evaluations pipeline using KID metric to compare the generated images with the real ones. The best value recorded from this evaluation is presented in Table \ref{tab1}.
 
Although Exp 10 achieved the lowest value (the best) in quantitative evaluation with the metric KID, and Exp 7 obtained the second lowest (second best) KID; the images in Exp 7 appear visually more realistic  than those in Exp 10. To explore this further, FID metric is calculated, to cross-validate the results by two experiments. The results are reported in Table\ref{tab3}. As shown in Table \ref{tab3}, the best KID and FID values are obtained from Exp 7 and Exp 10 at different stages of the training process. In Exp 7, the best value obtained by both metrics are at the training step 80; on the other hand, in Exp 10, the best value obtained by KID is at the training step 160 and the best value obtained by FID is at the step 80. Therefore, it is challenging to measure the realism of Rosacea artifact generated on the images based on these quantitative evaluations, specifically for Exp 10. Hence, the images obtained from both experiments needed visual scrutiny to check the fidelity of Rosacea.
 
From visual scrutiny, the generated images from Exp 7 were evaluated of higher fidelity than the images obtained from Exp 10. As discussed in Section \ref{sec:Truncation Trick}, and shown in Figure \ref{fig7}, the generated images from Exp 10 are blurred and lack variation in Rosacea features. As a result, the images generated from Exp 10 were not included in the further analysis.  

As mentioned in Section \ref{sec:Qualitative Evaluation}, the images obtained from Exp 7 were verified by the experts (dermatologists). Based on to the dermatologists' opinions, 73\% of the images got more than 60\% mean score, and the dermatologists remarks are provided in Table \ref{tab2}. Based on the non-specialist participants' opinions, 80\% of the images got more than 60\% mean score. In a nutshell, the StyleGAN2-ADA with the experimental fine-tuning described earlier in the paper produced high-quality realistic results as confirmed by experts and non-specialists participants. 
 
Based on these quantitative and qualitative evaluations, it is conceivable that metrics such as KID and FID are not sufficient by themselves as evaluation criteria when working with a limited dataset of medical images. Both quantitative and qualitative evaluations of the synthetic images demonstrate that, although the evaluation metrics such as FID, IS, and KID are used widely, they have many limitations to be aware of while working with medical images. Along with the quantitative evaluation, the qualitative assessment, such as expert opinion, may well be requisite in the computer-aided medical diagnosis community.

\begin{table}
\centering
	\caption{Dermatologists comments on generated Rosacea faces}
	%\label{table}
	%\setlength{\tabcolsep}{3pt}
	\begin{tabular}{|p{30pt}|p{40pt}|p{40pt}|p{40pt}|p{40pt}|}
		\hline
		Exp no. & Top KID value achieved & at step no. & Top FID value achieved & at step no.\\ \hline
		7 & 3.1 & 80 & 31.67 & 80 \\ \hline
		10 & 2.5 & 160 & 31.40 & 80 \\ \hline
	\end{tabular}
	\label{tab3}
\end{table}

\section{Future work}
\label{sec:Future work}
Given the importance of the hyperparameter $\gamma$ as discussed earlier, it would be desirable to design an Adaptive Regularization Technique \cite{zhao2019learning} for the weight matrix to be experimentally tested for StyleGAN2 architectures. Designing an adaptive $\gamma$ value would not only help in generating high fidelity synthetic images but can help achieve the equilibrium at the early stages during the training with limited samples. Reaching the equilibrium point at the earlier stages may help in reducing training time and cost without compromising the quality in the output. 

Adding this adaptive technique for $\gamma$ may also help in optimizing the model by introducing an automated early stopping point to the training process as it starts to overfit. This may save unnecessary time and cost, while the training is still under progress even after overfitting. 

In future, the generated images could be used to expand the dataset for classification of Rosacea. As the fidelity of the generated images improves, they could be used for Rosacea awareness, education, and advertisement purposes for the disease. Along with Rosacea, more facial diseases could be included for the same. 

As discussed in section \ref{sec:Limitations and Discussion}, popular metrics such as IS, FID, KID, Perpetual Path Length, Precision and Recall should not be considered as the only metrics in the assessment pipeline of synthetic medical images. However, it is necessary to have a quantitative evaluation to navigate the results/outputs by GAN models; hence it is essential to explore and improve the quantitative evaluation methods that may be deemed appropriate for the medical imaging domain. To achieve this, it is crucial to understand the nature of medical imaging with respect to imaging modality, fidelity and how to retain domain-specific information in synthetic images.

\section{Conclusion}
\label{sec:Conclusion}
In this study, we have demonstrated the effectiveness of using StyleGAN2-ADA to generate high-quality synthetic images of Rosacea from a small dataset of only 300 real images. By controlling the $R_1$ regularization weight, we were able to achieve this result, which serves as foundational work for investigating the use of advanced generative models in synthetic data generation for medical imaging with limited data. The conducted experiments also revealed that granular details of the skin disease can be generated by working with hyperparameters such as $R_1$ regularization, applying a limited set of augmentation techniques such as 'pixel blitting' and 'colour' and the Freeze-D technique with Transfer Learning. A qualitative analysis was conducted, in which expert dermatologists evaluated the generated images of Rosacea, and the mean opinion score indicated that 73\% of the generated images present a realistic pattern of Rosacea. Additionally, this study suggests that metrics such as KID and FID may have limitations in evaluating synthetic images generated from small datasets in the medical and clinical imaging field. The generated images were also evaluated by non-expert participants, which shows the synthetic Rosacea faces look realistic as 80\% of the images in the study have achieved the mean score 60\% and more.

\section*{Acknowledgments}
This work was conducted with the financial support of the Science Foundation Ireland Centre for Research Training in Digitally-Enhanced Reality (d-real) under Grant No. 18/CRT/6224.

%Bibliography
\bibliographystyle{unsrt}  
%\bibliography{refs}  

\end{document}